\documentclass{article}

\usepackage[final]{neurips_2021}

\usepackage[utf8]{inputenc} 
\usepackage[T1]{fontenc}    
\usepackage[pagebackref=true,
      breaklinks=true,
      letterpaper=true,
      colorlinks = true,
      linkcolor = red,
      urlcolor  = blue,
      citecolor = green,
      anchorcolor = blue]{hyperref}
\usepackage{url}            
\usepackage{booktabs}       
\usepackage{amsfonts}       
\usepackage{nicefrac}       
\usepackage{microtype}      
\usepackage{xcolor}         
\usepackage{graphicx}
\usepackage{amsmath}
\usepackage{bm}
\usepackage{multirow}
\usepackage{tabu}
\usepackage{siunitx}
\usepackage{adjustbox}
\usepackage{subcaption}
\usepackage{siunitx}
\usepackage{xcolor}
\usepackage{colortbl}
\usepackage{color}
\usepackage{subcaption}
\usepackage{pifont}
\definecolor{Gray}{gray}{0.95}
\definecolor{Cyan}{rgb}{0.88,1,1}

\usepackage{tabu}           
\usepackage{multirow}
\usepackage{booktabs}
\usepackage{makecell}

\newcolumntype{x}[1]{>{\centering\arraybackslash}p{#1pt}}
\newlength\savewidth

\newcommand{\tablestyle}[2]{\setlength{\tabcolsep}{#1}\renewcommand{\arraystretch}{#2}\centering\footnotesize}

\title{DynamicViT: Efficient Vision Transformers with Dynamic Token Sparsification}

\author{
Yongming Rao$^1$ 
~~
Wenliang Zhao$^1$  
~~
Benlin Liu$^{2,3}$ \\[0.1cm]
\textbf{Jiwen Lu$^1$}\thanks{Corresponding author.}
~~~
\textbf{Jie Zhou$^1$}
~~
\textbf{Cho-Jui Hsieh$^2$} 
~~
\\ [0.25cm]
$^1$ Tsinghua University ~~~~   $^2$ UCLA ~~~~ $^3$ University of Washington
}

\newcommand{\dynamvit}{DynamicViT}
\newcommand\cb[1]{\color{blue} #1}

\DeclareMathOperator{\MLP}{MLP}
\DeclareMathOperator{\Agg}{Agg}
\DeclareMathOperator{\softmax}{Softmax}
\DeclareMathOperator{\gumbelsoftmax}{Gumbel-Softmax}
\DeclareMathOperator{\kld}{KL}
\DeclareMathOperator{\CE}{CrossEntropy}
\DeclareMathOperator{\argsort}{argsort}

\begin{document}

\maketitle

\begin{abstract}
  Attention is sparse in vision transformers. We observe the final prediction in vision transformers is only based on a subset of most informative tokens, which is sufficient for accurate image recognition. Based on this observation, we propose a dynamic token sparsification framework to prune redundant tokens progressively and dynamically based on the input. Specifically, we devise a lightweight prediction module to estimate the importance score of each token given the current features. The module is added to different layers to prune redundant tokens hierarchically. To optimize the prediction module in an end-to-end manner, we propose an attention masking strategy to differentiably prune a token by blocking its interactions with other tokens. Benefiting from the nature of self-attention, the unstructured sparse tokens are still hardware friendly, which makes our framework easy to achieve actual speed-up. By hierarchically pruning 66\% of the input tokens, our method greatly reduces 31\% $\sim$ 37\%  FLOPs and improves the throughput by over 40\% while the drop of accuracy is within 0.5\% for various vision transformers. Equipped with the dynamic token sparsification framework,  DynamicViT models can achieve very competitive complexity/accuracy trade-offs compared to state-of-the-art CNNs and vision transformers on ImageNet. Code is available at \url{https://github.com/raoyongming/DynamicViT}.
\end{abstract}

\section{Introduction}

These years have witnessed the great progress in computer vision brought by the evolution of CNN-type architectures~\cite{he2016deep,krizhevsky2012alex}. Some recent works start to replace CNN by using transformer for many vision tasks, like object detection~\cite{zhu2020deformable,liu2021swin} and classification~\cite{touvron2020deit}. Just like what has been done to the CNN-type architectures in the past few years, it is also desirable to accelerate the transformer-like models to make them more suitable for real-time applications.

\begin{figure}[t]
    \centering
    \includegraphics[width=\textwidth]{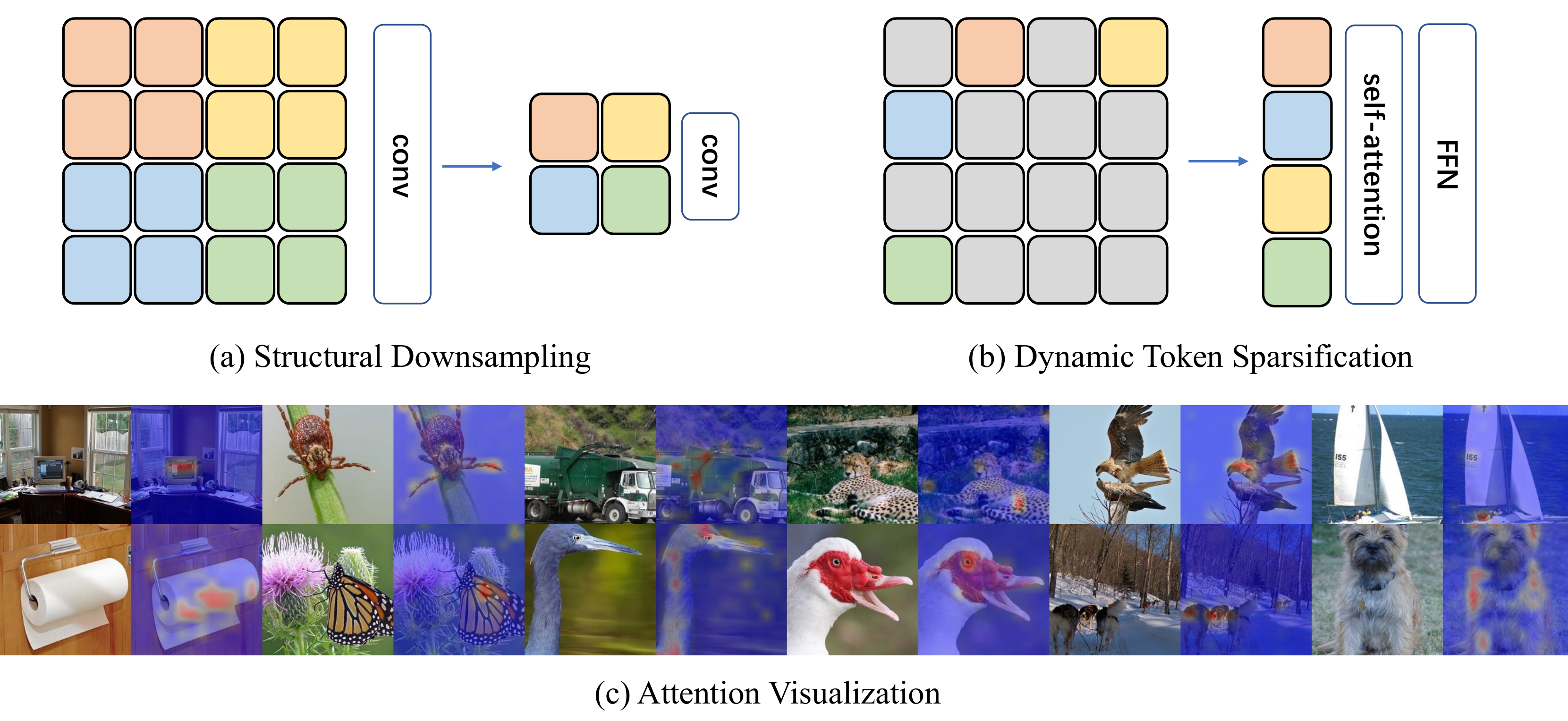}
    \caption{\textbf{Illustration of our main idea.} CNN models usually leverage the structural downsampling strategy to build hierarchical architectures as shown in (a). \emph{unstructured} and \emph{data-dependent} downsampling method in (b) can better exploit the sparsity in the input data. Thanks to the nature of the self-attention operation, the unstructured token set is also easy to accelerate through parallel computing. (c) visualizes the impact of each spatial location on the final prediction in the DeiT-S model~\cite{touvron2020deit} using the visualization method proposed in~\cite{chefer2020transformer}. These results demonstrate the final prediction in vision transformers is only based on a subset of most informative tokens, which suggests a large proportion of tokens can be removed without hurting the performance.} \vspace{-10pt}
    \label{fig:idea}
\end{figure}

One common practice for the acceleration of CNN-type networks is to prune the filters that are of less importance. The way input is processed by the vision transformer and its variants, \ie splitting the input image into multiple independent patches, provides us another orthogonal way to introduce the sparsity for the acceleration. That is, we can prune the tokens of less importance in the input instance, given the fact that many tokens contribute very little to the final prediction.  This is only possible for the transformer-like models where the self-attention module can take the token sequence of variable length as input, and the unstructured pruned input will not affect the self-attention module, while dropping a certain part of the pixels can not really accelerate the convolution operation since the unstructured neighborhood used by convolution would make it difficult to accelerate through parallel computing. Since the hierarchical architecture of CNNs with structural downsampling has improved model efficiency in various vision tasks, we hope to explore the \emph{unstructured} and \emph{data-dependent} downsampling strategy for vision transformers to further leverage the advantages of self-attention (our experiments also show unstructured sparsification can lead to better performance for vision transformers compared to structural downsampling). The basic idea of our method is illustrated in Figure~\ref{fig:idea}.

In this work, we propose to employ a lightweight prediction module to determine which tokens to be pruned in a dynamic way, dubbed as DynamicViT. In particular, for each input instance, the prediction module produces a customized binary decision mask to decide which tokens are uninformative and need to be abandoned. This module is added to multiple layers of the vision transformer, such that the sparsification can be performed in a hierarchical way as we gradually increase the amount of pruned tokens after each prediction module. Once a token is pruned after a certain layer, it will not be ever used in the feed-forward procedure. 
The additional computational overhead introduced by this lightweight module is quite small, especially considering the computational overhead saved by eliminating the uninformative tokens.

This prediction module can be optimized jointly in an end-to-end manner together with the vision transformer backbone. To this end, two specialized strategies are adopted. The first one is to adopt Gumbel-Softmax~\cite{eric2017gumbel} to overcome the non-differentiable problem of sampling from a distribution so that it is possible to perform the end-to-end training. The second one is about how to apply this learned binary decision mask to prune the unnecessary tokens. Considering the number of zero elements in the binary decision mask is different for each instance, directly eliminating the uninformative tokens for each input instance during training will make parallel computing impossible. Moreover, this would also hinder the back-propagation for the prediction module, which needs to calculate the probability distribution of whether to keep the token even if it is finally eliminated. Besides, directly setting the abandoned tokens as zero vectors is also not a wise idea since zero vectors will still affect the calculation of the attention matrix. Therefore, we propose a strategy called attention masking where we drop the connection from abandoned tokens to all other tokens in the attention matrix based on the binary decision mask. By doing so, we can overcome the difficulties described above. We also modify the original training objective of the vision transformer by adding a term to constrain the proportion of pruned tokens after a certain layer. During the inference phase, we can directly abandon a fixed amount of tokens after certain layers for each input instance as we no longer need to consider whether the operation is differentiable, and this will greatly accelerate the inference.

We illustrate the effectiveness of our method on ImageNet using DeiT~\cite{touvron2020deit} and LV-ViT~\cite{jiang2021token} as backbone. The experimental results demonstrate the competitive trade-off between speed and accuracy. In particular, by hierarchically pruning 66\% of the input tokens, we can greatly reduce 31\% $\sim$ 37\% GFLOPs and improve the throughput by over 40\% while the drop of accuracy is within 0.5\% for all different vision transformers. Our DynamicViT demonstrates the possibility of exploiting the sparsity in space for the acceleration of transformer-like model. We expect our attempt to open a new path for future work on the acceleration of transformer-like models.

\section{Related Work}
\paragraph{Vision transformers. } Transformer model is first widely studied in NLP community~\cite{vaswani2017attention}. It proves the possibility to use self-attention to replace the recurrent neural networks and their variants. Recent progress has demonstrated the variants of transformers can also be a competitive alternative to CNNs and achieve promising results on different vision tasks including image classification~\cite{dosovitskiy2020vit,touvron2020deit,liu2021swin,zhou2021deepvit,rao2021global}, object detection~\cite{carion2020end}, semantic segmentation~\cite{SETR,cheng2021maskformer} and 3D analysis~\cite{yu2021pointr,zhao2020point}. DETR~\cite{carion2020end} is the first work to apply the transformer model to vision tasks. It formulates the object detection task as a set prediction problem and follows the encoder-decoder design in the transformer to generate a sequence of bounding boxes. ViT~\cite{dosovitskiy2020vit} is the first work to directly apply transformer architecture on non-overlapping image patches for the image classification task, and the whole framework contains no convolution operation. Compared to CNN-type models, ViT can achieve better performance with large-scale pre-training. It is really preferred if the architecture can achieve the state-of-the-art without any pre-training. DeiT~\cite{touvron2020deit} proposes many training techniques so that we can train the convolution-free transformer only on ImageNet1K~\cite{deng2009imagenet} and achieve better performance than ViT. LV-ViT~\cite{jiang2021token} further improves the performance by introducing a new training objective called token labeling.  Both ViT and its follow-ups split the input image into multiple independent image patches and transform these image patches into tokens for further process. This makes it feasible to incorporate the sparsity in space dimension for these transformer-like models. 

\paragraph{Model acceleration. }
Model acceleration techniques are important for the deployment of deep models on edge devices. There are many techniques can be used to accelerate the inference speed of deep model, including quantization~\cite{gong2014compressing, wang2019haq}, pruning~\cite{he2017channel, rao2018runtime}, low-rank factorization~\cite{yu2017compressing}, knowledge distillation~\cite{hinton2015distilling, liu2020metadistiller} and so on. There are also many works aims at accelerating the inference speed of transformer models. For example, TinyBERT~\cite{jiao2019tinybert} proposes a distillation method to accelerate the inference of transformer. Star-Transformer~\cite{guo2019star} reduces quadratic space and time complexity to linear by replacing the fully connected structure with a star-shaped topology. However, all these works focus on NLP tasks, and few works explore the possibility of making use of the characteristic of vision tasks to accelerate vision transformer. Furthermore, the difference between the characteristics of Transformer and CNN also makes it possible to adopt another way for acceleration rather than the methods used for CNN acceleration like filter pruning~\cite{he2017channel}, which removes non-critical or redundant neurons from a deep model. Our method aims at pruning the tokens of less importance instead of the neurons by exploiting the sparsity of informative image patches.

\section{Dynamic Vision Transformers}

\begin{figure}[t]
    \centering
    \includegraphics[width=0.8\textwidth]{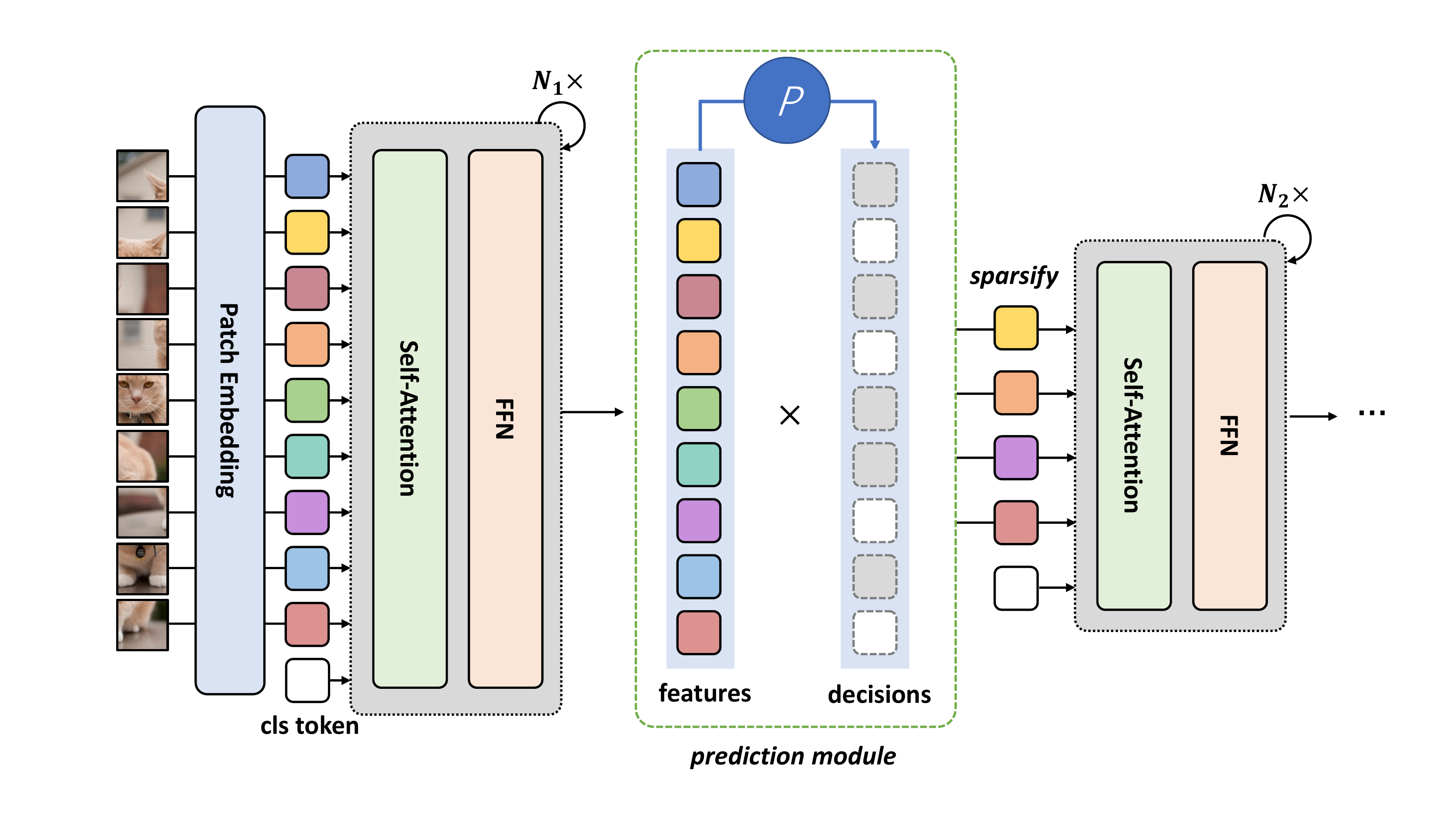}
    \caption{\textbf{The overall framework of the proposed approach.} The proposed prediction module is inserted between the transformer blocks to selectively prune less informative token conditioned on features produced by the previous layer. By doing so, less tokens are processed in the followed layers. }
    \label{fig:overall}
\end{figure}

\subsection{Overview}
The overall framework of our \dynamvit~is illustrated in Figure~\ref{fig:overall}. Our \dynamvit{} consists of a normal vision transformer as the backbone and several prediction modules. The backbone network can be implemented as a wide range of vision transformer (\eg, ViT~\cite{dosovitskiy2020vit}, DeiT~\cite{touvron2020deit}, LV-ViT~\cite{jiang2021token}). The prediction modules are responsible for generating the probabilities of dropping/keeping the tokens. The token sparsification is performed hierarchically through the whole network at certain locations. For example, given a 12-layer transformer, we can conduct token sparsification before the 4th, 7th, and 10th blocks. During training, the prediction modules and the backbone network can be optimized in an end-to-end manner thanks to our newly devised attention masking strategy. During inference, we only need to select the most informative tokens according to a predefined pruning ratio and the scores computed by the prediction modules. 

\subsection{Hierarchical Token Sparsification with Prediction Modules}
An important characteristic of our \dynamvit{} is that the token sparsification is performed hierarchically, \ie, we gradually drop the uninformative tokens as the computation proceeds. To achieve this, we maintain a binary decision mask $\hat{\mathbf{D}}\in \{0, 1\}^{N}$ to indicate whether to drop or keep each token, where $N=HW$ is the number of patch embeddings\footnote{We omit the class token for simplicity, while in practice we always keep the class token (\ie, the decision for class token is always ``1'').}. We initialize all elements in the decision mask to 1 and update the mask progressively. The prediction modules take the current decision $\hat{\mathbf{D}}$ and the tokens $\mathbf{x}\in\mathbb{R}^{N\times C}$ as input. We first project the tokens using an MLP:
\begin{equation}
    \mathbf{z}^{\rm local} = \MLP(\mathbf{x})\in \mathbb{R}^{N\times C'},
\end{equation}
where $C'$ can be a smaller dimension and we use $C'=C/2$ in our implementation. Similarly, we can compute a global feature by:
\begin{equation}
    \mathbf{z}^{\rm global}=\Agg(\MLP(\mathbf{x}), \hat{\mathbf{D}})\in \mathbb{R}^{C'},
\end{equation}
where $\Agg$ is the function which aggregate the information all the existing tokens and can be simply implemented as an average pooling:
\begin{equation}
    \Agg(\mathbf{u}, \hat{\mathbf{D}}) = \frac{\sum_{i=1}^{N}\hat{\mathbf{D}}_i \mathbf{u}_i}{\sum_{i=1}^{N}\hat{\mathbf{D}}_i}, \quad \mathbf{u}\in \mathbb{R}^{N\times C'}.
\end{equation}
The local feature encodes the information of a certain token while the global feature contains the context of the whole image, thus both of them are informative. Therefore, we combine both the local and global features to obtain local-global embeddings and feed them to another MLP to predict the probabilities to drop/keep the tokens:
\begin{align}
    &\mathbf{z}_i = [\mathbf{z}_i^{\rm local}, \mathbf{z}_i^{\rm global}],\quad 1\le i\le N,\\
    &\bm{\pi} = \softmax(\MLP(\mathbf{z})) \in \mathbb{R}^{N\times 2},
\end{align}
where $\bm{\pi}_{i, 0}$ denotes the probability of dropping the $i$-th token and $\bm{\pi}_{i, 1}$ is the probability of keeping it. We can then generate current decision $\mathbf{D}$ by sampling from $\bm{\pi}$ and update $\hat{\mathbf{D}}$ by
\begin{equation}
    \hat{\mathbf{D}}\leftarrow \hat{\mathbf{D}} \odot \mathbf{D},
\end{equation}
where $\odot$ is the Hadamard product, indicating that once a token is dropped, it will never be used.


\subsection{End-to-end Optimization with Attention Masking}
Although our target is to perform token sparsification, we find it non-trivial to implement in practice during training. First, the sampling from $\bm{\pi}$ to get binary decision mask $\mathbf{D}$ is is non-differentiable, which impedes the end-to-end training. To overcome this, we  apply the Gumbel-Softmax technique~\cite{eric2017gumbel} to sample from the probabilities $\bm{\pi}$:
\begin{equation}
    \mathbf{D} = \gumbelsoftmax(\bm{\pi})_{*, 1}\in \{0, 1\}^{N},
\end{equation}
where we use the index ``1'' because $\mathbf{D}$ represents the mask of the \emph{kept} tokens. The output of  Gumbel-Softmax is a
one-hot tensor, of which the expectation equals $\bm{\pi}$ exactly. Meanwhile, Gumbel-Softmax is differentiable thus makes it possible for end-to-end training.

The second obstacle comes when we try to prune the tokens during training. The decision mask $\hat{\mathbf{D}}$ is usually unstructured and the masks for different samples contain various numbers of 1's. Therefore, simply discarding the tokens where $\hat{\mathbf{D}}_i=0$ would result in a non-uniform number of tokens for samples within a batch,
which makes it hard to parallelize the computation.
Thus, we must keep the number of tokens unchanged, while cut down the interactions between the pruned tokens and other tokens. We also find that merely zero-out the tokens to be dropped using the binary mask $\hat{\mathbf{D}}$ is not feasible, because in the calculation of self-attention matrix~\cite{vaswani2017attention}
\begin{equation}
    \mathbf{A} = \softmax\left(\frac{\mathbf{Q}\mathbf{K}^T}{\sqrt{C}}\right)
\end{equation}
the zeroed tokens will still influence other tokens through the $\softmax$ operation. To this end, we devise a strategy called attention masking which can totally eliminate the effects of the dropped tokens. Specifically, we compute the attention matrix by:
\begin{align}
    &\mathbf{P} = \mathbf{Q}\mathbf{K}^T/\sqrt{C} \in \mathbb{R}^{N\times N},\\
    &\mathbf{G}_{ij} = 
    \begin{cases}
    1,& i=j,\\
    \hat{\mathbf{D}}_j,& i\neq j.
    \end{cases}& 1\le i,j\le N,\label{equ:G_ij}\\
    &\tilde{\mathbf{A}}_{ij} =  \frac{\exp(\mathbf{P}_{ij})\mathbf{G}_{ij}}{\sum_{k=1}^N\exp(\mathbf{P}_{ik})\mathbf{G}_{ik}},& 1\le i,j\le N.\label{equ:attention_masking}
\end{align}
By Equation~\eqref{equ:G_ij} we construct a graph where $\mathbf{G}_{ij}=1$ means the $j$-th token will contribute to the update of the $i$-th token. Note that we explicitly add a self-loop to each token to improve numerically stability. It is also easy to show the self-loop does not influence the results: if $\hat{\mathbf{D}}_j=0$, the $j$-th token will not contribute to any tokens other than itself. Equation~\eqref{equ:attention_masking} computes the masked attention matrix $\tilde{\mathbf{A}}$, which is equivalent to the attention matrix calculated by considering only the kept tokens but has a constant shape $N\times N$ during training.

\subsection{Training and Inference}
We now describe the training objectives of our \dynamvit{}. The training of \dynamvit{} includes training the prediction modules such that they can produce favorable decisions and fine-tuning the backbone to make it adapt to token sparsification. Assuming we are dealing with a minibatch of $B$ samples, we adopt the standard cross-entropy loss:
\begin{equation}
    \mathcal{L}_{\rm cls} = \CE(\mathbf{y}, \bar{\mathbf{y}}),
\end{equation}
where $\mathbf{y}$ is the prediction of the \dynamvit{} (after softmax) and $\bar{\mathbf{y}}$ is the ground truth. 

To minimize the influence on performance caused by our token sparsification, we use the original backbone network as a teacher model and hope the behavior of our \dynamvit{} as close to the teacher model as possible. Specifically, we consider this constraint from two aspects. First, we make the finally remaining tokens of the \dynamvit{} close to the ones of the teacher model, which can be viewed as a kind of self-distillation:
\begin{equation}
    \mathcal{L}_{\rm distill} = \frac{1}{\sum_{b=1}^B\sum_{i=1}^N \hat{\mathbf{D}}_i^{b, S}}\sum_{b=1}^B\sum_{i=1}^N \hat{\mathbf{D}}_i^{b, S} (\mathbf{t}_{i} - \mathbf{t}_i')^2,
\end{equation}
where $\mathbf{t}_{i}$ and $\mathbf{t}_i'$ denotes the $i$-th token after the last block of the \dynamvit{} and the teacher model, respectively. $\hat{\mathbf{D}}^{b, s}$ is the decision mask for the $b$-th sample at the $s$-th sparsification stage. Second, we minimize the difference of the predictions between our \dynamvit{} and its teacher via the KL divergence:
\begin{equation}
    \mathcal{L}_{\rm KL} = \kld\left(\mathbf{y}\| \mathbf{y}'\right),
\end{equation}
where $\mathbf{y}'$ is the prediction of the teacher model. 

Finally, we want to constrain the ratio of the kept tokens to a predefined value. Given a set of target ratios for $S$ stages $\bm{\rho}=[\rho^{(1)}, \ldots, \rho^{(S)}]$, we utilize an MSE loss to supervise the prediction module:
\begin{equation}
    \mathcal{L}_{\rm ratio} = \frac{1}{BS}\sum_{b=1}^{B}\sum_{s=1}^S \left(\rho^{(s)} - \frac{1}{N}\sum_{i=1}^N\hat{\mathbf{D}}^{b, s}_i\right)^2.
\end{equation}

The full training objective is a combination of the above objectives:
\begin{equation}
    \mathcal{L} = \mathcal{L}_{\rm cls} + \lambda_{\rm KL}\mathcal{L}_{\rm KL} +  \lambda_{\rm distill}\mathcal{L}_{\rm distill} + \lambda_{\rm ratio}\mathcal{L}_{\rm ratio},
\end{equation}
where we set $ \lambda_{\rm KL}=0.5,\lambda_{\rm distill}=0.5,\lambda_{\rm ratio}=2$ in all our experiments.

During inference, given the target ratio $\bm{\rho}$, we can directly discard the less informative tokens via the probabilities produced by the prediction modules such that only exact $m^s=\lfloor\rho^sN\rfloor$ tokens are kept at the $s$-th stage. Formally, for the $s$-th stage, let
\begin{equation}
\mathcal{I}^s = \argsort(\bm{\pi}_{*, 1})
\end{equation}
be the indices sorted by the keeping probabilities $\bm{\pi}_{*, 1}$, we can then keep the tokens of which the indices lie in $\mathcal{I}^s_{1:m^s}$ while discarding the others. In this way, our \dynamvit{} prunes less informative tokens dynamically at runtime, thus can reduce the computational costs during inference.

\section{Experimental Results}\label{sec:experiemnt}
\newcommand{\imnetacc}{ImageNet Acc. (\%)}
\newcommand{\throughput}{Throughput (im/s)}
\begin{table}[t]
  \centering
  \caption{\textbf{Main results on ImageNet.} We apply our method on three representative vision transformers: DeiT-S, LV-ViT-S and LV-ViT-M. DeiT-S~\cite{touvron2020deit} is a widely used vision transformer with the simple architecture. LV-ViT-S and LV-ViT-M~\cite{jiang2021token} are the state-of-the-art vision transformers. We report the top-1 classification accuracy, theoretical complexity in FLOPs and throughput for different ratio $\rho$. The throughput is measured on a single NVIDIA RTX 3090 GPU with batch size fixed to 32. }
  \adjustbox{width=\textwidth}{
    \begin{tabu}to 1.16\textwidth{ll*{4}{X[l]}}\toprule
    \multirow{2}[0]{*}{Base Model} & \multicolumn{1}{c}{\multirow{2}[0]{*}{Metrics}} & \multicolumn{4}{c}{Keeping Ratio $\rho$ at each stage} \\\cmidrule{3-6}
          &       & 1.0   & 0.9   & 0.8   & 0.7  \\\midrule
    \multirow{3}[0]{*}{DeiT-S~\cite{touvron2020deit}} & \imnetacc{}  & 79.8  & 79.8 {\cb(-0.0)}  & 79.6 {\cb(-0.2)}  & 79.3 {\cb(-0.5)} \\
          & GFLOPs & 4.6   & 4.0 {\cb(-14\%)}   & 3.4  {\cb(-27\%)}   & 2.9  {\cb(-37\%)}  \\
          & \throughput{} & 1337.7  & 1524.8  {\cb(+14\%)}  & 1774.6  {\cb(+33\%)}  & 2062.1  {\cb(+54\%)}  \\\midrule
    \multirow{3}[0]{*}{LV-ViT-S~\cite{jiang2021token}} & \imnetacc{}  & 83.3  & 83.3 {\cb(-0.0)}  & 83.2 {\cb(-0.1)}  & 83.0 {\cb(-0.3)}  \\
          & GFLOPs & 6.6   & 5.8 {\cb(-12\%)}    & 5.1 {\cb(-22\%)}    & 4.6 {\cb(-31\%)}   \\
          & \throughput{} & 993.3  & 1108.3 {\cb(+12\%)}  & 1255.6 {\cb(+26\%)}   & 1417.6 {\cb(+43\%)}   \\\midrule
    \multirow{3}[0]{*}{LV-ViT-M~\cite{jiang2021token}} & \imnetacc{}  & 84.0  & 83.9 {\cb(-0.1)}  & 83.9 {\cb(-0.1)}  & 83.8 {\cb(-0.2)}  \\
          & GFLOPs & 12.7  & 11.1 {\cb(-13\%)}  & 9.6 {\cb(-24\%)}    & 8.5 {\cb(-33\%)}   \\
          & \throughput{} & 589.5  & 688.5 {\cb(+17\%)}   & 791.2 {\cb(+34\%)}  & 888.2 {\cb(+50\%)}  \\\bottomrule
    \end{tabu}%
}
  \label{tab:main}%
\end{table}%
In this section, we will demonstrate the superiority of the proposed \dynamvit{} through extensive experiments. In all of our experiments, we fix the number of sparsification stages $S=3$ and apply the target keeping ratio $\bm{\rho}$ as a geometric sequence $[\rho, \rho^2, \rho^3]$ where $\rho$ ranges from $(0, 1)$. During training DynamicViT models, we follow most of the training techniques used in DeiT~\cite{touvron2020deit}. We use the pre-trained vision transformer models to initialize the backbone models  and jointly train the whole model for 30 epochs. We set the learning rate of the prediction module to $\frac{\text{batch size}}{1024}\times 0.001$ and use $0.01\times$ learning rate for the backbone model. We fix the weights of the backbone models in the first 5 epochs. All of our models are trained on a single machine with 8 GPUs. Other training setups and details can be found in the supplementary material.

\subsection{Main results} One of the most advantages of the~\dynamvit{} is that it can be applied to a wide range of vision transformer architectures to reduce the computational complexity with minor loss of performance. In Table~\ref{tab:main}, we summarize the main results on ImageNet~\cite{deng2009imagenet} where we evaluate our~\dynamvit{} used three base models (DeiT-S~\cite{touvron2020deit}, LV-ViT-S~\cite{jiang2021token} and LV-ViT-M~\cite{jiang2021token}). We report the top-1 accuracy, FLOPs, and the throughput under different keeping ratios $\rho$. Note that our token sparsification is performed hierarchically in three stages, there are only $\lfloor N\rho^3\rfloor$ tokens left after the last stage. The throughput is measured on a single NVIDIA RTX 3090 GPU with batch size fixed to 32. We demonstrate that our \dynamvit{} can reduce the computational costs by $31\%\sim 37\%$ and accelerate the inference at runtime by $43\%\sim 54\%$, with the neglectable influence of performance ($-0.2\%\sim -0.5\%$).

\subsection{Comparisons with the-state-of-the-arts}
\begin{figure}[t]
\centering
\begin{minipage}{0.48\textwidth}
\centering
\includegraphics[height=6cm]{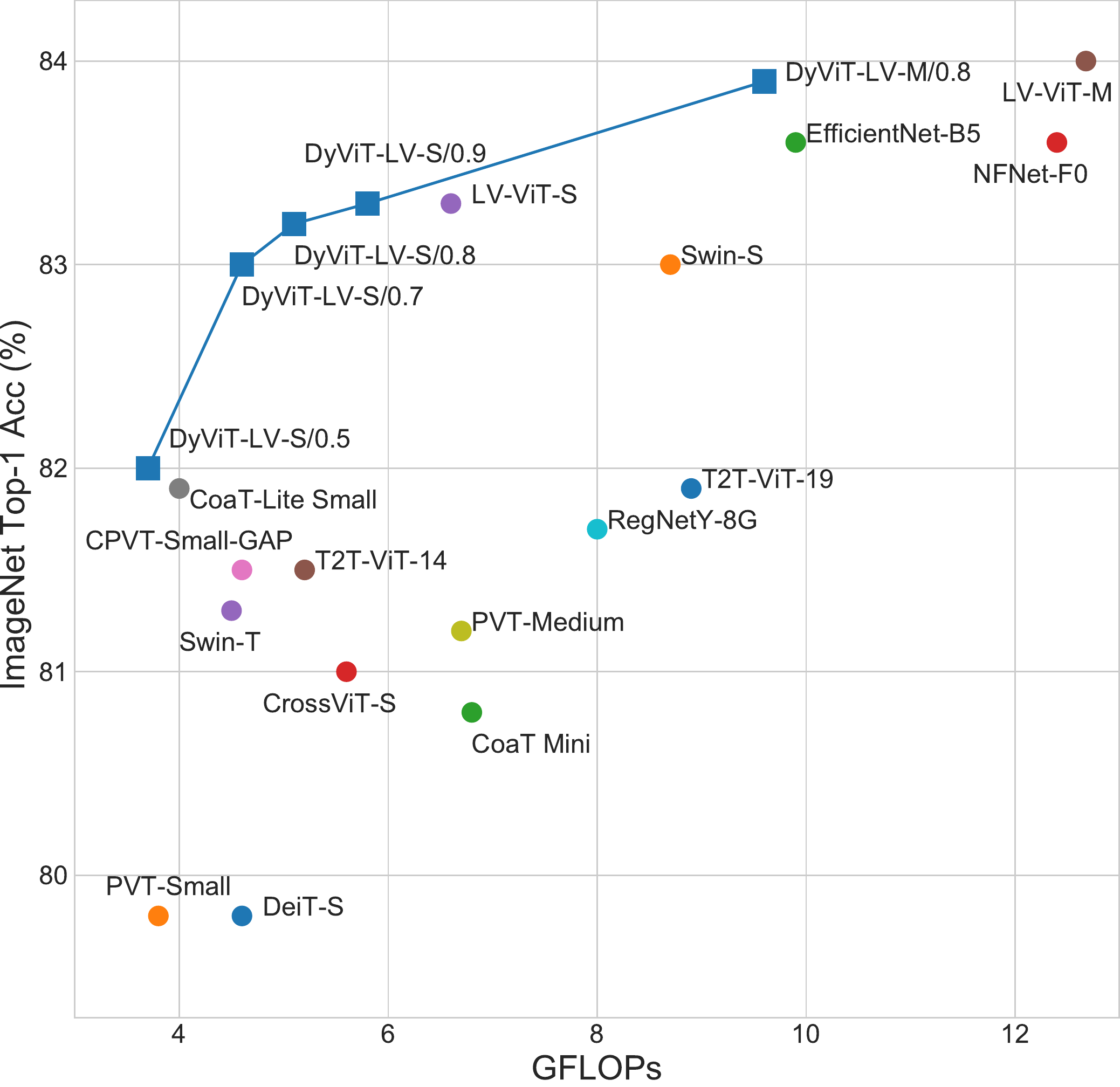}
\caption{\small Model complexity (FLOPs) and top-1 accuracy trade-offs on ImageNet. We compare DynamicViT with the state-of-the-art image classification models. Our models achieve better trade-offs compared to the various vision transformers as well as carefully designed CNN models. }\label{fig:sota_flops_acc}
\end{minipage}\hfill
\begin{minipage}{0.48\textwidth}
\centering
\includegraphics[height=6cm]{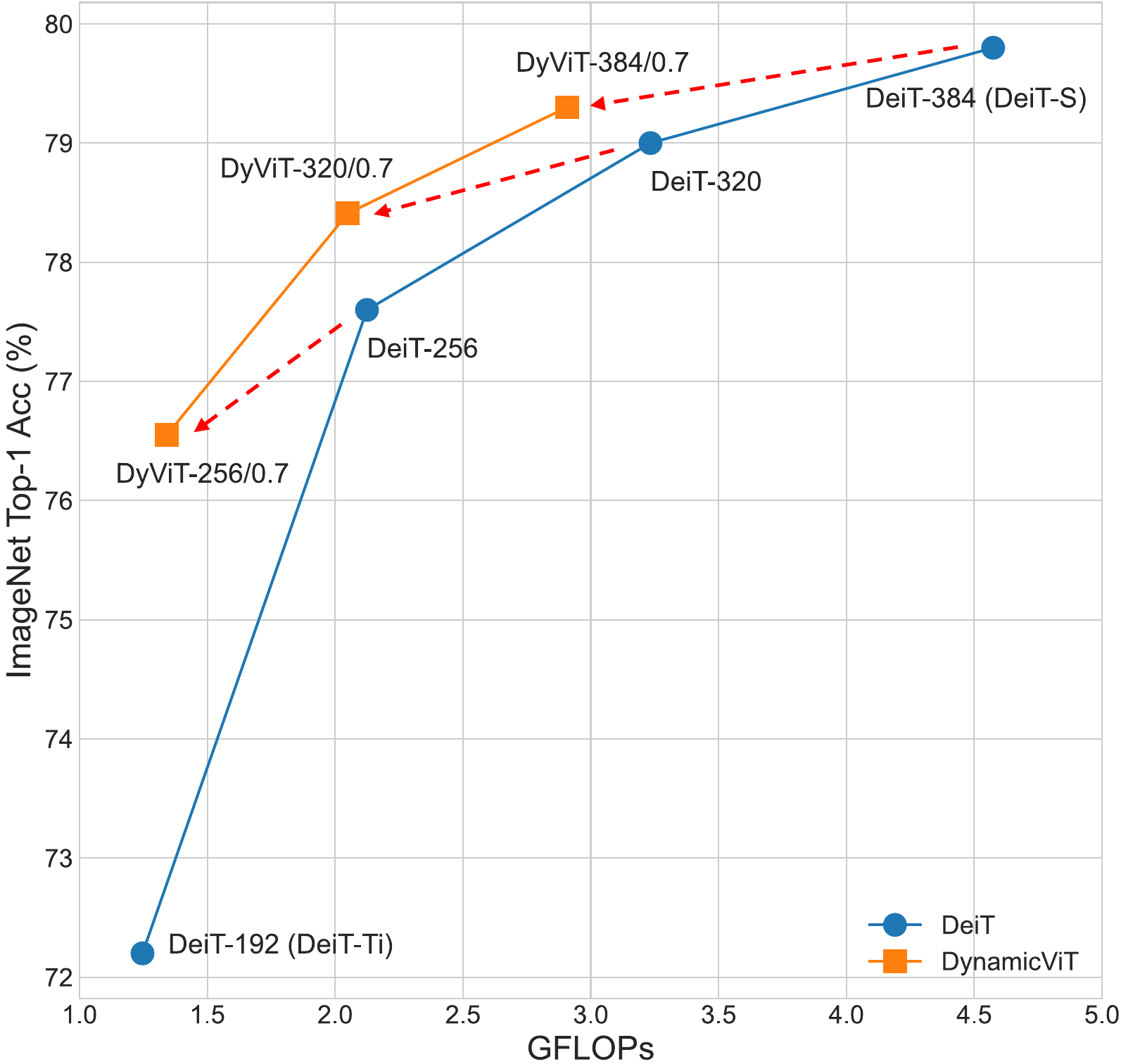}
\caption{\small Comparison of our dynamic token sparsification method with model width scaling. We train our \dynamvit{} based on DeiT models with embedding dimension varying from 192 to 384 and fix ratio $\rho=0.7$. We see dynamic token sparsification is more efficient than commonly used  model width scaling.}\label{fig:deit_acc}
\end{minipage}
\end{figure}

In Table~\ref{tab:sota}, we compare the~\dynamvit{} with the state-of-the-art models in image classification, including convolutional networks and transformer-like architectures. We use the \dynamvit{} with LV-ViT~\cite{jiang2021token} as the base model and use the ``$/\rho$'' to indicate the keeping ratio. We observe that our \dynamvit{} exhibits favorable complexity/accuracy trade-offs at all three complexity levels. Notably, we find our DynamicViT-LV-M/0.7 beats the EfficientNet-B5~\cite{tan2019efficientnet} and NFNet-F0~\cite{brock2021nfnet}, which are two of the current state-of-the-arts CNN architectures. This can also be shown clearer in Figure~\ref{fig:sota_flops_acc}, where we plot the FLOPS-accuracy curve of \dynamvit{} series  (where we use DyViT for short), along with  other state-of-the-art models. We can also observe that \dynamvit{} can achieve better trade-offs than LV-ViT series, which strongly demonstrates the effectiveness of our method.

\begin{table}[t]
  \centering
  \caption{\textbf{Comparisons with the state-of-the-arts on ImageNet.} We compare our DynamicViT models with state-of-the-art image classifciation models with comparable FLOPs and number of parameters. We use the \dynamvit{} with LV-ViT~\cite{jiang2021token} as the base model and use the ``$/\rho$'' to indicate the keeping ratio. We also include the results of LV-ViT models as references. }
    \begin{tabu}to\textwidth{l*{4}{X[c]}}\toprule
    Model & Params (M) & GFLOPs & Resolution   & Top-1 Acc (\%) \\\midrule
    DeiT-S~\cite{touvron2020deit} & 22.1  & 4.6   & 224   & 79.8  \\
    PVT-Small~\cite{wang2021pvt} & 24.5  & 3.8   & 224   & 79.8  \\
    CoaT Mini~\cite{xu2021coat} & 10.0  & 6.8   & 224   & 80.8  \\
    CrossViT-S~\cite{chen2021crossvit} & 26.7  & 5.6   & 224   & 81.0  \\
    PVT-Medium~\cite{wang2021pvt} & 44.2  & 6.7   & 224   & 81.2  \\
    Swin-T~\cite{liu2021swin} & 29.0  & 4.5   & 766   & 81.3  \\
    T2T-ViT-14~\cite{yuan2021t2t} & 22.0  & 5.2   & 224   & 81.5  \\
    CPVT-Small-GAP~\cite{chu2021cpvt} & 23.0  & 4.6   & 817   & 81.5  \\
    CoaT-Lite Small~\cite{xu2021coat} & 20.0  & 4.0   & 224   & 81.9  \\\midrule
    LV-ViT-S~\cite{jiang2021token} & 26.2 & 6.6 & 224 & 83.3 \\
    DynamicViT-LV-S/0.5 & 26.9  & 3.7   & 224   & 82.0  \\
    DynamicViT-LV-S/0.7 & 26.9  & 4.6   & 224   & 83.0  \\\midrule \midrule
    RegNetY-8G~\cite{radosavovic2020designing} & 39.0  & 8.0   & 224   & 81.7  \\
    T2T-ViT-19~\cite{yuan2021t2t} & 39.2  & 8.9   & 224   & 81.9  \\
    Swin-S~\cite{liu2021swin} & 50.0  & 8.7   & 224   & 83.0  \\
    EfficientNet-B5~\cite{tan2019efficientnet} & 30.0  & 9.9   & 456   & 83.6  \\
    NFNet-F0~\cite{brock2021nfnet} & 72.0  & 12.4  & 256   & 83.6  \\\midrule
    DynamicViT-LV-M/0.7 & 57.1  & 8.5   & 224   & 83.8  \\\midrule \midrule
    ViT-Base/16~\cite{dosovitskiy2020vit} & 86.6  & 17.6  & 224   & 77.9  \\
    DeiT-Base/16~\cite{touvron2020deit} & 86.6  & 17.6  & 224   & 81.8  \\
    CrossViT-B~\cite{chen2021crossvit} & 104.7  & 21.2  & 224   & 82.2  \\
    T2T-ViT-24~\cite{yuan2021t2t} & 64.1  & 14.1  & 224   & 82.3  \\
    TNT-B~\cite{han2021transformer} & 66.0  & 14.1  & 224   & 82.8  \\
    RegNetY-16G~\cite{radosavovic2020designing} & 84.0  & 16.0  & 224   & 82.9  \\
    Swin-B~\cite{liu2021swin} & 88.0  & 15.4  & 224   & 83.3  \\\midrule
    LV-ViT-M~\cite{jiang2021token} & 55.8 & 12.7 & 224 & 84.0 \\
    DynamicViT-LV-M/0.8 & 57.1  & 9.6   & 224   & 83.9  \\\bottomrule
    \end{tabu}%
  \label{tab:sota}%
\end{table}%

\subsection{Analysis}
\paragraph{\dynamvit{} for model scaling.} The success of EfficientNet~\cite{tan2019efficientnet} shows that we can obtain a model with better complexity/accuracy tradeoffs by scaling the model along different dimensions. While in vision transformers, the most commonly used method to scale the model is to change the number of channels, our \dynamvit{} provides another powerful tool to perform token sparsification. We analysis this nice property of \dynamvit{} in Figure~\ref{fig:deit_acc}. First, we train several DeiT~\cite{touvron2020deit} models with the embedding dimension varying from 192 (DeiT-Ti) to 384 (DeiT-S). Second, we train our \dynamvit{} based on those models with the keeping ratio $\rho=0.7$. We find that after performing token sparsification, the complexity of the model is reduced to be similar to its variant with a smaller embedding dimension. Specifically, we observe that by applying our \dynamvit{} to DeiT-256, we obtain a model that has a comparable computational complexity to DeiT-Ti, but enjoys around $4.3\%$ higher ImageNet top-1 accuracy.

\begin{figure}
    \centering
    \includegraphics[width=\textwidth]{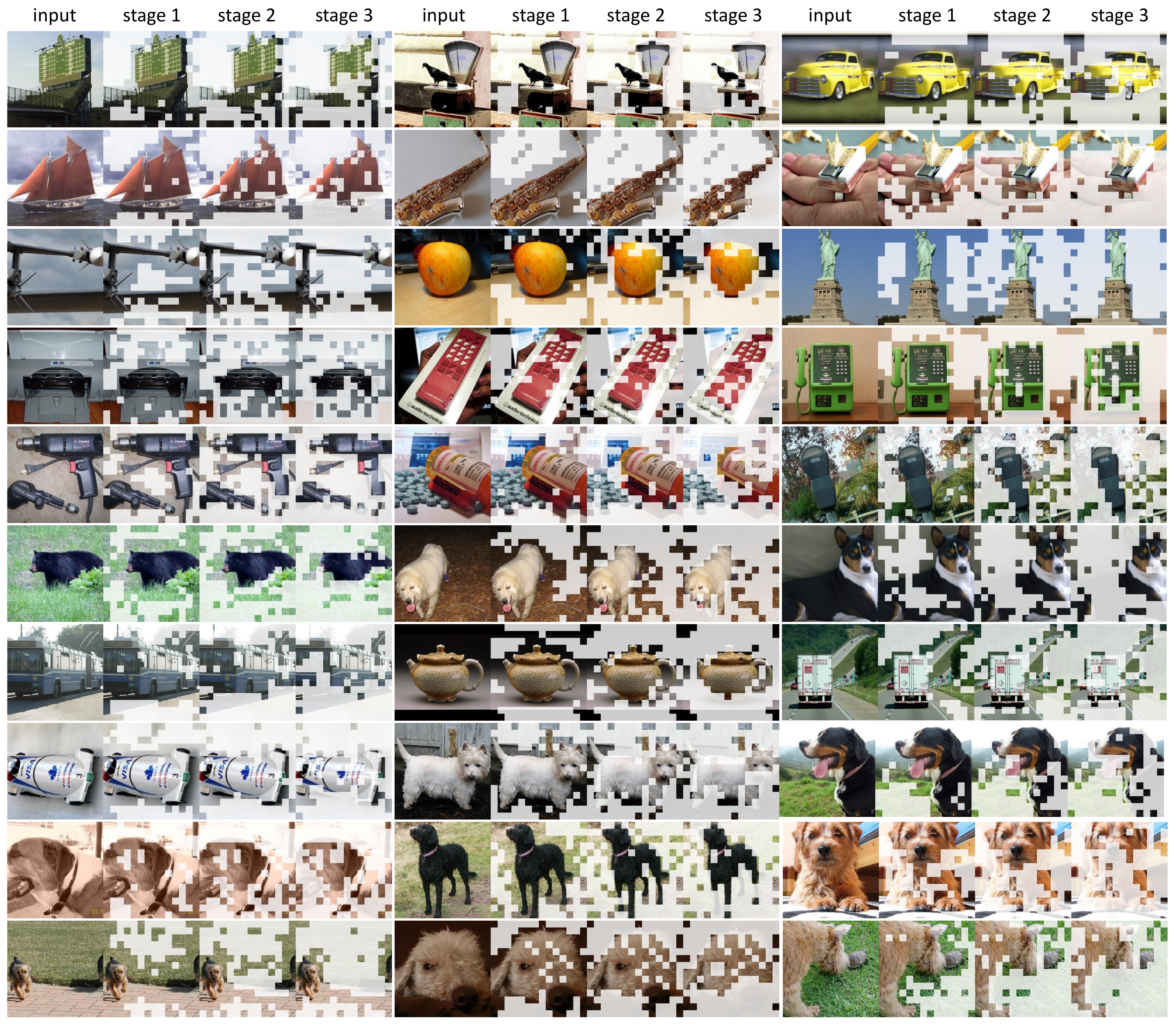}
    \caption{\textbf{Visualization of the progressively sparsified tokens.}  We show the original input image and the sparsification results after the three stages, where the masks represent the corresponding tokens are discarded. We see our method can gradually focus on the most representative regions in the image.  This phenomenon suggests that the~\dynamvit{} has better interpretability. }
    \label{fig:viz}
\end{figure}

\paragraph{Visualizations.} To further investigate the behavior of \dynamvit{}, we visualize the sparsification procedure in Figure~\ref{fig:viz}. We show the original input image and the sparsification results after the three stages, where the masks represent the corresponding tokens are discarded. We find that through the hierarchically token sparsification, our \dynamvit{} can gradually drop the uninformative tokens and finally focus on the objects in the images. This phenomenon also suggests that the~\dynamvit{} leads to  better interpretability, \ie, it can locate the important parts in the image which contribute most to the classification step-by-step.

\begin{figure}[t]
\centering
\begin{minipage}{0.4\textwidth}
\captionsetup{type=table}
\caption{\small Effects of different losses. We provide the results after removing the distillation loss and the KL loss.}
\label{tab:loss}
\adjustbox{width=\linewidth}{
    \begin{tabu}to 1.3\linewidth{lXX}\toprule
    Base Model&\multicolumn{1}{l}{DeiT-S}&\multicolumn{1}{l}{LVViT-S}\\\midrule
    DynamicViT&79.3\cb{(-0.5)}&83.0\cb{(-0.3)}\\
    w/o distill (Eq.13)&79.3\cb{(-0.5)}&82.7\cb{(-0.6)}\\
    w/o KL (Eq.14)&79.2\cb{(-0.6)}&82.9\cb{(-0.4)}\\
    w/o distill \& KL&79.2\cb{(-0.6)}&82.5\cb{(-0.8)}\\\bottomrule
    \end{tabu}%
    }
\end{minipage}\hfill
\begin{minipage}{0.6\textwidth}
\centering
\includegraphics[height=2.6cm]{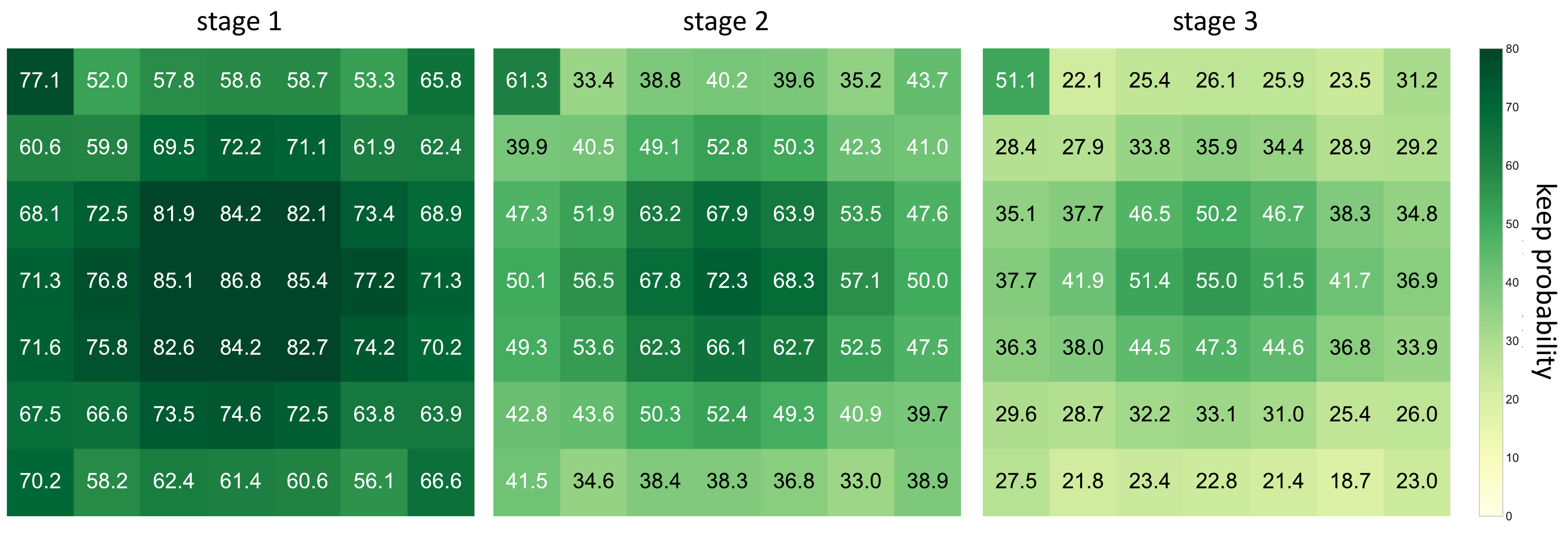}
\caption{\small The keep probabilities of the tokens at each stage.}\label{fig:keep_probl}
\end{minipage}
\end{figure}

Besides the sample-wise visualization we have shown above, we are also interested in the statistical characteristics of the sparsification decisions, \ie, what kind of general patterns does the \dynamvit{} learn from the dataset? We then use the \dynamvit{} to generate the decisions for all the images in the ImageNet validation set and compute the keep probability of each token in all three stages, as shown in Figure~\ref{fig:keep_probl}. We average pool the probability maps into $7\times 7$ such that they can be visualized more easily. Unsurprisingly, we find the tokens in the middle of the image tend to be kept, which is reasonable because in most images the objects are located in the center. We can also find that the later stage generally has lower probabilities to be kept, mainly because that the keeping ratio at the $s$ stage is $\rho^s$, which decreases exponentially as $s$ increases.

\begin{table}[t]
\caption{Comparisons of different sparsification strategies. We investigate different methods to select redundant tokens based on the DeiT-S model. We report the top-1 accuracy on ImageNet for different methods. We fix the complexity of the accelerated models to 2.9G FLOPs for fair comparisons. } \label{tab:ablation}
\centering
\begin{subtable}{.32\textwidth}
\caption{\small Dynamic sparsification \vs static/structural downsampling.}
\adjustbox{height=1cm}{
 \begin{tabu}to \linewidth{X[l]c}\toprule
Model & \multicolumn{1}{l}{Acc. (\%)}  \\\midrule
Structural & 78.2~\cb{(-1.6)} \\
Static & 73.4~\cb{(-6.4)}  \\
\rowcolor{Gray} Dynamic & 79.3~\cb{(-0.5)} \\\bottomrule
\end{tabu}%
}
\end{subtable}\hfill
\begin{subtable}{.32\textwidth}
\caption{\small Different redundant token removal methods.}
\adjustbox{height=1cm}{
\begin{tabu}to    \linewidth{X[l]c}
\toprule
Model &  Acc. (\%)\\ \midrule
Random & 77.5 \cb{(-2.3)} \\ 
Attention & 78.1 \cb{(-1.7)}\\
\rowcolor{Gray} Prediction & 79.3 \cb{(-0.5)} \\\bottomrule
\end{tabu}
}
\end{subtable}\hfill
\begin{subtable}{.32\textwidth}
\caption{\small Effects of number of sparsification stages.}
\adjustbox{height=1cm}{
  \begin{tabu}to \linewidth{X[l]c}\toprule
    Model  & Acc. (\%)  \\\midrule
    Single-stage & 77.4~\cb{(-2.4)} \\
    Two-stage & 79.2~\cb{(-0.6)}  \\
    \rowcolor{Gray} Three-stage & 79.3~\cb{(-0.5)} \\\bottomrule
\end{tabu}
}

\end{subtable}
\end{table}

\begin{table}[t]
\caption{Results on larger models. We apply our method to the model with larger width (\ie, DeiT-B) and the model with larger input size (\ie, DeiT-S with $384\times 384$ input). } \label{tab:large}
\centering
\subfloat[\small Results on DeiT-B.]
{\makebox[0.48\linewidth][c]{
\tablestyle{12pt}{1.2}
\adjustbox{width=0.48\linewidth}{
\setlength{\tabcolsep}{7pt}
\begin{tabular}{lll}
    \toprule
        Model & GFLOPs & Acc. (\%) \\ \midrule
        DeiT-B & 17.5 & 81.8 \\
        DynamicViT-B/0.7 & 11.2 \cb{(-36\%)} & 81.3 \cb{(-0.5)} \\ \bottomrule
    \end{tabular}
\label{Tab:ablation:a}}
} 
}
\hfill
\subfloat[\small Results on the $384\times384$ input.]
{\makebox[0.48\linewidth][c]{
\tablestyle{12pt}{1.2}
\adjustbox{width=0.48\linewidth}{
\setlength{\tabcolsep}{7pt}
\begin{tabular}{lll}
    \toprule
        Model & GFLOPs & Acc. (\%) \\ \midrule
        DeiT-S & 15.5 & 81.6 \\ 
        DynamicViT-S/0.7 & 9.5 \cb{(-39\%)} & 81.4 \cb{(-0.2)} \\
        DynamicViT-S/0.5 & 7.0 \cb{(-55\%)} & 80.3 \cb{(-1.3)} \\ \bottomrule
    \end{tabular}
}
}
}
\vspace{-5pt}
\end{table}

\paragraph{Effects of different losses.} We show the effects of different losses in Table~\ref{tab:loss}.  We see the improvement brought by the distillation loss and the KL loss is not very significant, but it can consistently further boost the performance of various models.

\paragraph{Comparisons of different sparsification strategies.} As illustrated in Figure~\ref{fig:overall}, the dynamic token sparsification is unstructured. To discuss whether the dynamic sparsification is better than other strategies, we perform ablation experiments and the results are shown in Table~\ref{tab:ablation}. For the structural downsampling, we perform an average pooling with kernel size $2\times 2$ after the sixth block of the baseline DeiT-S~\cite{touvron2020deit} model, which has similar FLOPs to our DynamicViT. The static token sparsification means that the sparsification decisions are not conditioned on the input tokens. We also compare our method with other token removal methods like randomly removing tokens or removing tokens based the attention score of the class token. We find through the experiments that although other strategies have similar computational complexities, the proposed dynamic token sparsification method achieves the best accuracy. We also show that the progressive sparsification method is significantly better than one-stage sparsification.

\paragraph{Accelerating larger models.} To show the effectiveness of our method on larger models, we apply our method to the model with larger width (\ie, DeiT-B) and models with larger input size (\ie, DeiT-S with $384\times 384$ input). The results are presented in Table~\ref{tab:large}. We see our method also works well on the larger DeiT model. The accuracy drop become less significant when we apply our method to the model with larger feature maps. Notably, we can reduce the complexity of the DeiT-S model with $384\times 384$ input by over 50\% with only 1.3\% accuracy drop.

\section{Conclusion}
\label{sec:con}
In this work, we open a new path to accelerate vision transformer by exploiting the sparsity of informative patches in the input image. For each input instance, our DynamicViT model prunes the tokens of less importance in a dynamic way according to the customized binary decision mask output from the lightweight prediction module, which fuses the local and global information containing in the tokens. The prediction module is added to multiple layers such that the token pruning is performed in a hierarchical way. Gumbel-Softmax and attention masking techniques are also incorporated for the end-to-end training of the transformer model together with the prediction module. During the inference phase, our approach can greatly improves the efficiency by gradually pruning 66\% of the input tokens, while the drop of accuracy is less than 0.5\% for different transformer backbone. In this paper, we focus on the image classification task. Extending our method to other scenarios like video classification and dense prediction tasks can be interesting directions.

\subsection*{Acknowledgment}

This work was supported in part by the National Key Research and Development Program of China under Grant 2017YFA0700802, in part by the National Natural Science Foundation of China under Grant 62125603, Grant 61822603, Grant U1813218, Grant U1713214, in part by Beijing Academy of Artificial Intelligence (BAAI), in part by National Science Foundation  under grant IIS-1901527, IIS-2008173, IIS-2048280, 
and in part by a grant from the Institute for Guo Qiang, Tsinghua University.

\bibliographystyle{plain}
\bibliography{ref}

\begin{thebibliography}{10}

\bibitem{brock2021nfnet}
Andrew Brock, Soham De, Samuel~L Smith, and Karen Simonyan.
\newblock High-performance large-scale image recognition without normalization.
\newblock {\em arXiv preprint arXiv:2102.06171}, 2021.

\bibitem{carion2020end}
Nicolas Carion, Francisco Massa, Gabriel Synnaeve, Nicolas Usunier, Alexander
  Kirillov, and Sergey Zagoruyko.
\newblock End-to-end object detection with transformers.
\newblock In {\em ECCV}, pages 213--229, 2020.

\bibitem{chefer2020transformer}
Hila Chefer, Shir Gur, and Lior Wolf.
\newblock Transformer interpretability beyond attention visualization.
\newblock {\em arXiv preprint arXiv:2012.09838}, 2020.

\bibitem{chen2021crossvit}
Chun-Fu Chen, Quanfu Fan, and Rameswar Panda.
\newblock Crossvit: Cross-attention multi-scale vision transformer for image
  classification.
\newblock {\em arXiv preprint arXiv:2103.14899}, 2021.

\bibitem{cheng2021maskformer}
Bowen Cheng, Alexander~G. Schwing, and Alexander Kirillov.
\newblock Per-pixel classification is not all you need for semantic
  segmentation.
\newblock {\em NeurIPS}, 2021.

\bibitem{chu2021cpvt}
Xiangxiang Chu, Zhi Tian, Bo~Zhang, Xinlong Wang, Xiaolin Wei, Huaxia Xia, and
  Chunhua Shen.
\newblock Conditional positional encodings for vision transformers.
\newblock {\em arXiv preprint arXiv:2102.10882}, 2021.

\bibitem{deng2009imagenet}
Jia Deng, Wei Dong, Richard Socher, Li-Jia Li, Kai Li, and Li~Fei-Fei.
\newblock Imagenet: A large-scale hierarchical image database.
\newblock In {\em CVPR}, pages 248--255, 2009.

\bibitem{dosovitskiy2020vit}
Alexey Dosovitskiy, Lucas Beyer, Alexander Kolesnikov, Dirk Weissenborn,
  Xiaohua Zhai, Thomas Unterthiner, Mostafa Dehghani, Matthias Minderer, Georg
  Heigold, Sylvain Gelly, Jakob Uszkoreit, and Neil Houlsby.
\newblock An image is worth 16x16 words: Transformers for image recognition at
  scale.
\newblock {\em arXiv preprint arXiv:2010.11929}, 2020.

\bibitem{gong2014compressing}
Yunchao Gong, Liu Liu, Ming Yang, and Lubomir Bourdev.
\newblock Compressing deep convolutional networks using vector quantization.
\newblock {\em arXiv preprint arXiv:1412.6115}, 2014.

\bibitem{guo2019star}
Qipeng Guo, Xipeng Qiu, Pengfei Liu, Yunfan Shao, Xiangyang Xue, and Zheng
  Zhang.
\newblock Star-transformer.
\newblock {\em arXiv preprint arXiv:1902.09113}, 2019.

\bibitem{han2021transformer}
Kai Han, An~Xiao, Enhua Wu, Jianyuan Guo, Chunjing Xu, and Yunhe Wang.
\newblock Transformer in transformer.
\newblock {\em arXiv preprint arXiv:2103.00112}, 2021.

\bibitem{he2016deep}
Kaiming He, Xiangyu Zhang, Shaoqing Ren, and Jian Sun.
\newblock Deep residual learning for image recognition.
\newblock In {\em CVPR}, pages 770--778, 2016.

\bibitem{he2017channel}
Yihui He, Xiangyu Zhang, and Jian Sun.
\newblock Channel pruning for accelerating very deep neural networks.
\newblock In {\em ICCV}, pages 1389--1397, 2017.

\bibitem{hinton2015distilling}
Geoffrey Hinton, Oriol Vinyals, and Jeff Dean.
\newblock Distilling the knowledge in a neural network.
\newblock {\em arXiv preprint arXiv:1503.02531}, 2015.

\bibitem{eric2017gumbel}
Eric Jang, Shixiang Gu, and Ben Poole.
\newblock Categorical reparameterization with gumbel-softmax.
\newblock In {\em ICLR}, 2017.

\bibitem{jiang2021token}
Zihang Jiang, Qibin Hou, Li~Yuan, Daquan Zhou, Xiaojie Jin, Anran Wang, and
  Jiashi Feng.
\newblock Token labeling: Training a 85.5\% top-1 accuracy vision transformer
  with 56m parameters on imagenet.
\newblock {\em arXiv preprint arXiv:2104.10858}, 2021.

\bibitem{jiao2019tinybert}
Xiaoqi Jiao, Yichun Yin, Lifeng Shang, Xin Jiang, Xiao Chen, Linlin Li, Fang
  Wang, and Qun Liu.
\newblock Tinybert: Distilling bert for natural language understanding.
\newblock {\em arXiv preprint arXiv:1909.10351}, 2019.

\bibitem{krizhevsky2012alex}
Alex Krizhevsky, Ilya Sutskever, and Geoffrey~E Hinton.
\newblock Imagenet classification with deep convolutional neural networks.
\newblock {\em NeurIPS}, 25:1097--1105, 2012.

\bibitem{liu2020metadistiller}
Benlin Liu, Yongming Rao, Jiwen Lu, Jie Zhou, and Cho-Jui Hsieh.
\newblock Metadistiller: Network self-boosting via meta-learned top-down
  distillation.
\newblock In {\em European Conference on Computer Vision}, pages 694--709.
  Springer, 2020.

\bibitem{liu2021swin}
Ze~Liu, Yutong Lin, Yue Cao, Han Hu, Yixuan Wei, Zheng Zhang, Stephen Lin, and
  Baining Guo.
\newblock Swin transformer: Hierarchical vision transformer using shifted
  windows.
\newblock {\em arXiv preprint arXiv:2103.14030}, 2021.

\bibitem{radosavovic2020designing}
Ilija Radosavovic, Raj~Prateek Kosaraju, Ross Girshick, Kaiming He, and Piotr
  Doll{\'a}r.
\newblock Designing network design spaces.
\newblock In {\em CVPR}, pages 10428--10436, 2020.

\bibitem{rao2018runtime}
Yongming Rao, Jiwen Lu, Ji~Lin, and Jie Zhou.
\newblock Runtime network routing for efficient image classification.
\newblock {\em IEEE transactions on pattern analysis and machine intelligence},
  41(10):2291--2304, 2018.

\bibitem{rao2021global}
Yongming Rao, Wenliang Zhao, Zheng Zhu, Jiwen Lu, and Jie Zhou.
\newblock Global filter networks for image classification.
\newblock In {\em NeurIPS}, 2021.

\bibitem{tan2019efficientnet}
Mingxing Tan and Quoc Le.
\newblock Efficientnet: Rethinking model scaling for convolutional neural
  networks.
\newblock In {\em ICML}, pages 6105--6114. PMLR, 2019.

\bibitem{touvron2020deit}
Hugo Touvron, Matthieu Cord, Matthijs Douze, Francisco Massa, Alexandre
  Sablayrolles, and Herv{\'e} J{\'e}gou.
\newblock Training data-efficient image transformers \& distillation through
  attention.
\newblock {\em arXiv preprint arXiv:2012.12877}, 2020.

\bibitem{vaswani2017attention}
Ashish Vaswani, Noam Shazeer, Niki Parmar, Jakob Uszkoreit, Llion Jones,
  Aidan~N Gomez, Lukasz Kaiser, and Illia Polosukhin.
\newblock Attention is all you need.
\newblock {\em arXiv preprint arXiv:1706.03762}, 2017.

\bibitem{wang2019haq}
Kuan Wang, Zhijian Liu, Yujun Lin, Ji~Lin, and Song Han.
\newblock Haq: Hardware-aware automated quantization with mixed precision.
\newblock In {\em Proceedings of the IEEE/CVF Conference on Computer Vision and
  Pattern Recognition}, pages 8612--8620, 2019.

\bibitem{wang2021pvt}
Wenhai Wang, Enze Xie, Xiang Li, Deng-Ping Fan, Kaitao Song, Ding Liang, Tong
  Lu, Ping Luo, and Ling Shao.
\newblock Pyramid vision transformer: A versatile backbone for dense prediction
  without convolutions.
\newblock {\em arXiv preprint arXiv:2102.12122}, 2021.

\bibitem{xu2021coat}
Weijian Xu, Yifan Xu, Tyler Chang, and Zhuowen Tu.
\newblock Co-scale conv-attentional image transformers.
\newblock {\em arXiv preprint arXiv:2104.06399}, 2021.

\bibitem{yu2017compressing}
Xiyu Yu, Tongliang Liu, Xinchao Wang, and Dacheng Tao.
\newblock On compressing deep models by low rank and sparse decomposition.
\newblock In {\em CVPR}, pages 7370--7379, 2017.

\bibitem{yu2021pointr}
Xumin Yu, Yongming Rao, Ziyi Wang, Zuyan Liu, Jiwen Lu, and Jie Zhou.
\newblock Pointr: Diverse point cloud completion with geometry-aware
  transformers.
\newblock In {\em ICCV}, 2021.

\bibitem{yuan2021t2t}
Li~Yuan, Yunpeng Chen, Tao Wang, Weihao Yu, Yujun Shi, Zihang Jiang, Francis~EH
  Tay, Jiashi Feng, and Shuicheng Yan.
\newblock Tokens-to-token vit: Training vision transformers from scratch on
  imagenet.
\newblock {\em arXiv preprint arXiv:2101.11986}, 2021.

\bibitem{zhao2020point}
Hengshuang Zhao, Li~Jiang, Jiaya Jia, Philip Torr, and Vladlen Koltun.
\newblock Point transformer.
\newblock In {\em ICCV}, 2021.

\bibitem{SETR}
Sixiao Zheng, Jiachen Lu, Hengshuang Zhao, Xiatian Zhu, Zekun Luo, Yabiao Wang,
  Yanwei Fu, Jianfeng Feng, Tao Xiang, Philip~H.S. Torr, and Li~Zhang.
\newblock Rethinking semantic segmentation from a sequence-to-sequence
  perspective with transformers.
\newblock In {\em CVPR}, 2021.

\bibitem{zhou2021deepvit}
Daquan Zhou, Bingyi Kang, Xiaojie Jin, Linjie Yang, Xiaochen Lian, Qibin Hou,
  and Jiashi Feng.
\newblock Deepvit: Towards deeper vision transformer.
\newblock {\em arXiv preprint arXiv:2103.11886}, 2021.

\bibitem{zhu2020deformable}
Xizhou Zhu, Weijie Su, Lewei Lu, Bin Li, Xiaogang Wang, and Jifeng Dai.
\newblock Deformable detr: Deformable transformers for end-to-end object
  detection.
\newblock {\em arXiv preprint arXiv:2010.04159}, 2020.

\end{thebibliography}

\appendix

\section{Implementation Details}\label{sec:details}

We conduct our experiments on the ImageNet (also known as ILSVRC2012)~\cite{deng2009imagenet} dataset. ImageNet is a commonly used benchmark for image classification. We train our models on the training set, which consists of 1.28M images. The top-1 accuracy is measured on the 50k validation images following common practice~\cite{he2016deep,touvron2020deit}. To fairly compare with previous methods, we report the single crop results. 

We fix the number of sparsification stages $S=3$ in all of our experiments, since this setting can lead to a decent trade-off between complexity and performance. For the sake of simplicity, we set the target keeping ratio $\bm{\rho}$ as a geometric sequence $[\rho, \rho^2, \rho^3]$, where $\rho$ is the keeping ratio after each sparsifcation ranging from $(0, 1)$. For the prediction module, we use the identical architecture for different stages. We  use two \texttt{LayerNorm} $\to$ \texttt{Linear}($C$, $C/2$) $\to$ \texttt{GELU} block to produce $\mathbf{z}^{\rm local}$ and $ \mathbf{z}^{\rm global}$  respectively. We employ a \texttt{Linear}($C$, $C/2$) $\to$ \texttt{GELU} $\to$ \texttt{Linear}($C/2$, $C/4$) $\to$ \texttt{GELU} $\to$ \texttt{Linear}($C/4$, $2$) $\to$ \texttt{Softmax} block to predict the probabilities.

During training our DynamicViT models, we follow most of the training techniques used in DeiT~\cite{touvron2020deit}. We use the pre-trained vision transformer models to initialize the backbone models and jointly train the backbone model as well as the prediction modules for 30 epochs. We set the learning rate of the prediction module to $\frac{\text{batch size}}{1024}\times 0.001$ and use $0.01\times$  learning rate for the backbone model. The batch size is adjusted adaptively for different models according to the GPU memory.  We fix the weights of the backbone models in the first 5 epochs. All of our models can be trained on a single machine with 8 NVIDIA GTX 1080Ti GPUs.

\section{More Analysis}


In this section, we provide more analysis of our method. We investigate the effects of progressive sparsification, distillation loss, ratio loss, and keeping ratio. We also include more visualization results. The following describes the details of the experiments, results and analysis. 

\paragraph{Progressive sparsification. } To verify the effectiveness of the progressive sparsification strategy, we test different sparsification methods that result in similar overall complexity. Here we provide more detailed results and more analysis. We find that progressive sparsification is much better than single-shot sparsification. Increasing the number of stages will lead to better performance. Since further increasing the number of stages ($>3$) will not lead to significantly better performance but add computation, we use a 3-stage progressive sparsification strategy in our main experiments. 

\begin{table}[!h]
    \centering \small
    \newcolumntype{g}{>{\columncolor{Gray}}l}
   \begin{tabular}{gll}\toprule
     & \multicolumn{1}{l}{Top-1 accuracy (\%)} & \multicolumn{1}{l}{GFLOPs} \\\midrule
    DeiT-S~\cite{touvron2020deit} & 79.8  & 4.6 \\\midrule
    $\rho = 0.25, [\rho]$ (single-stage) & 77.4\cb{(-2.4)} & 2.9\cb{(-37\%)} \\
    $\rho = 0.60, [\rho, \rho^2]$ (two-stage) & 79.2\cb{(-0.6)}  & 2.9\cb{(-37\%)} \\
    $\rho = 0.70, [\rho, \rho^2, \rho^3]$ (three-stage) & 79.3\cb{(-0.5)}  & 2.9\cb{(-37\%)} \\\bottomrule
\end{tabular}
\end{table}

\paragraph{Ablation on the distillation loss and ratio loss. } The weights of the distillation losses and ratio loss are the key hyper-parameters in our method.  Since the token-wise distillation loss and the KL divergence loss play similar roles in our method, we set $ \lambda_{\rm KL} = \lambda_{\rm distill}$ in all of our experiments for the sake of simplicity. In this experiment, we fix the keeping ratio $\rho$ to be 0.7.  We find our method is not sensitive to these hyper-parameters in general. The proposed ratio loss can encourage the model to reach the desired acceleration rate.  Distillation losses can improve the performance after sparsification.  We directly apply the best hyper-parameters searched on DeiT-S for all models.

\begin{figure}[!h]
\centering
    \newcolumntype{g}{>{\columncolor{Gray}}l}
\begin{minipage}{0.48\textwidth} \centering
\captionsetup{type=table}
\small
\begin{tabular}{gl}\toprule
     & \multicolumn{1}{l}{Top-1 accuracy (\%)} \\\midrule
    DeiT-S~\cite{touvron2020deit} & 79.8  \\\midrule
    $ \lambda_{\rm KL} = \lambda_{\rm distill} = 0$ & 79.17\cb{(-0.63)}  \\
    $ \lambda_{\rm KL} = \lambda_{\rm distill} = 0.5$ & 79.32\cb{(-0.48)} \\
    $ \lambda_{\rm KL} = \lambda_{\rm distill} = 1 $ & 79.23\cb{(-0.57)} \\\bottomrule
\end{tabular}

\end{minipage}\hfill
\begin{minipage}{0.48\textwidth}
\centering
\small
\begin{tabular}{gl}\toprule
     & \multicolumn{1}{l}{Top-1 accuracy (\%)} \\\midrule
    DeiT-S~\cite{touvron2020deit} & 79.8  \\\midrule
    $ \lambda_{\rm ratio}= 1$ & 79.15\cb{(-0.65)}  \\
     $ \lambda_{\rm ratio}= 2$& 79.32\cb{(-0.48)} \\
     $ \lambda_{\rm ratio}= 4$ & 79.29\cb{(-0.51)} \\\bottomrule
\end{tabular}
\end{minipage}
\end{figure}

\paragraph{Smaller keeping ratio. } We have also tried applying a smaller keeping ratio (larger acceleration rate). The results based on DeiT-S~\cite{touvron2020deit} and LV-ViT-S~\cite{jiang2021token} models are presented in the following tables. We see that using $\rho < 0.7$ will lead to a significant accuracy drop while reducing fewer FLOPs. Since only 22\% and 13\% tokens are remaining in the last stage when we set $\rho$ to 0.6 and 0.5 respectively, small $\rho$ may cause a significant information loss. Therefore, we use $\rho \geq 0.7$ in our main experiments. Jointly scaling  $\rho$ and the model width can be a better solution to achieve a large acceleration rate as shown in Figure 4 in the paper.

\begin{figure}[!h]
\centering
    \newcolumntype{g}{>{\columncolor{Gray}}l}
\begin{minipage}{0.48\textwidth} \centering
\captionsetup{type=table}
\small
 \begin{tabular}{gll}\toprule
     & \multicolumn{1}{l}{Top-1 acc. (\%)} & \multicolumn{1}{l}{GFLOPs} \\\midrule
    DeiT-S~\cite{touvron2020deit} & 79.8  & 4.6 \\\midrule
    $\rho = 0.9$  & 79.8\cb{(-0.0)} & 4.0\cb{(-14\%)} \\
    $\rho = 0.8$ & 79.6\cb{(-0.3)}  & 3.4\cb{(-27\%)} \\
     $\rho = 0.7$  & 79.3\cb{(-0.5)} & 2.9\cb{(-37\%)} \\
    $\rho = 0.6$ & 78.5\cb{(-1.3)}  & 2.5\cb{(-46\%)} \\
    $\rho = 0.5$ & 77.5\cb{(-2.3)}  & 2.2\cb{(-52\%)} \\\bottomrule
\end{tabular}

\end{minipage}\hfill
\begin{minipage}{0.48\textwidth}
\centering
\small
 \begin{tabular}{gll}\toprule
     & \multicolumn{1}{l}{Top-1 acc. (\%)} & \multicolumn{1}{l}{GFLOPs} \\\midrule
    LV-ViT-S~\cite{jiang2021token} & 83.3  & 6.6 \\\midrule
    $\rho = 0.9$  & 83.3\cb{(-0.0)} & 5.8\cb{(-12\%)} \\
    $\rho = 0.8$ & 83.2\cb{(-0.1)}  & 5.1\cb{(-22\%)} \\
     $\rho = 0.7$  & 83.0\cb{(-0.3)} & 4.6\cb{(-31\%)} \\
    $\rho = 0.6$ & 82.6\cb{(-0.7)}  & 4.1\cb{(-38\%)} \\
    $\rho = 0.5$ & 82.0\cb{(-1.3)}  & 3.7\cb{(-44\%)} \\\bottomrule
\end{tabular}
\end{minipage}
\end{figure}

\begin{figure}
    \centering
    \includegraphics[width=\textwidth]{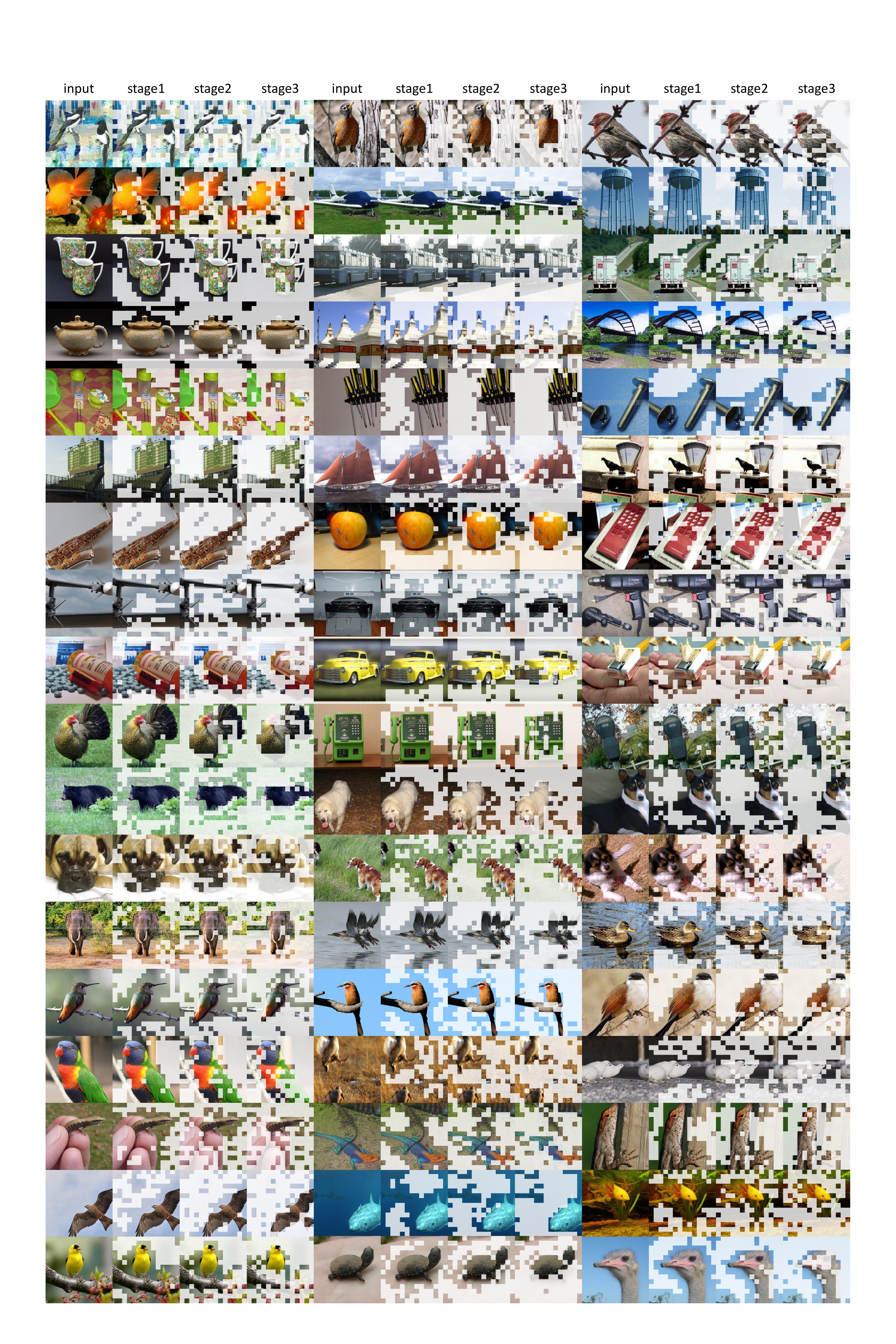}
    \caption{\textbf{More visual results}. The input images are randomly sampled from the validation set of ImageNet. We see our method works well for different images from various categories.}
    \label{fig:suppvis}
\end{figure}

\paragraph{More visual results. } We provide more visual results in Figure~\ref{fig:suppvis}. The input images are randomly sampled from the validation set of ImageNet. We see our method works well for different images from various categories.

\end{document}